\documentclass[manuscript,screen,review=false,authorversion=true,nonacm=true]{acmart}

\AtBeginDocument{%
  \providecommand\BibTeX{{%
    \normalfont B\kern-0.5em{\scshape i\kern-0.25em b}\kern-0.8em\TeX}}}

\acmJournal{TKDD}

\usepackage{graphicx}
\usepackage{multirow}
\usepackage{hhline}
\usepackage{colortbl}
\usepackage{xcolor}
\usepackage{booktabs}
\usepackage{bm}

\begin{document}

\title{Addressing practical challenges in Active Learning via a hybrid query strategy}

\author{Deepesh Agarwal}
\affiliation{%
  \institution{Kansas State University}
  \city{Mnahattan}
  \state{Kansas}
  \country{USA}
}
\email{deepesh@ksu.edu}

\author{Pravesh Srivastava}
\affiliation{%
  \institution{Indian Institute of Technology Gandhinagar}
  \city{Gandhinagar}
  \state{Gujarat}
  \country{India}
}
\email{srivastava_pravesh@alumni.iitgn.ac.in}

\author{Sergio Martin-del-Campo}
\affiliation{%
  \institution{Viking Analytics}
  \city{Gothenburg}
  \country{Sweden}
}
\affiliation{%
  \institution{Lule\aa \, University of Technology}
  \city{Lule\aa}
  \country{Sweden}
}
\email{sergio.mdcb@vikinganalytics.se}

\author{Balasubramaniam Natarajan}
\affiliation{%
  \institution{Kansas State University}
  \city{Mnahattan}
  \state{Kansas}
  \country{USA}
}
\email{bala@ksu.edu}

\author{Babji Srinivasan}
\affiliation{%
  \institution{Indian Institute of Technology Madras}
  \city{Chennai}
  \state{Tamil Nadu}
  \country{India}
}
\email{babji.srinivasan@iitm.ac.in}

\renewcommand{\shortauthors}{Agarwal, et al.}

\begin{abstract}
Active Learning (AL) is a powerful tool to address modern machine learning problems with significantly fewer labeled training instances. However, implementation of traditional AL methodologies in practical scenarios is accompanied by multiple challenges due to the inherent assumptions. There are several hindrances, such as unavailability of labels for the AL algorithm at the beginning; unreliable external source of labels during the querying process; or incompatible mechanisms to evaluate the performance of Active Learner. Inspired by these practical challenges, we present a hybrid query strategy-based AL framework that addresses three practical challenges simultaneously: cold-start, oracle uncertainty and performance evaluation of Active Learner in the absence of ground truth. While a pre-clustering approach is employed to address the cold-start problem, the uncertainty surrounding the expertise of labeler and confidence in the given labels is incorporated to handle oracle uncertainty. The heuristics obtained during the querying process serve as the fundamental premise for accessing the performance of Active Learner. The robustness of the proposed AL framework is evaluated across three different environments and industrial settings. The results demonstrate the capability of the proposed framework to tackle practical challenges during AL implementation in real-world scenarios. \footnote{This work has been submitted to the ACM for possible publication. Copyright may be transferred without notice, after which this version may no longer be accessible.}
\end{abstract}


\maketitle

\section{Introduction}

Active Learning (AL) constitutes an area of machine learning where the learning algorithm is allowed to interact with external sources of information to obtain labels corresponding to the unlabelled instances in the dataset. In these AL settings, the learning algorithm is termed as \textit{Active Learner} and the external source of information (i.e., a domain expert of any other labelling source) is referred to as the \textit{oracle}. The key idea is to achieve higher accuracy with fewer labelled instances when compared to conventional supervised learning methods. This is accomplished by allowing the Active Learner to choose the data it wants to learn from. Thus, the Active Learner poses \textit{queries} in the form of unlabelled data instances and requests the oracle to provide corresponding labels. The Active Learner uses the information acquired from the oracle to continuously update the trained model. AL methods are advantageous in modern machine learning frameworks, where the labelling process may be laborious, long-standing or expensive \cite{coldst1, tkdd_spec_01}. However, numerous challenges are encountered during implementation of AL frameworks in practical scenarios due to inherent assumptions \cite{settles2011theories}. This article presents a novel hybrid query strategy-based AL framework that addresses three practical challenges, namely, cold-start, oracle uncertainty and performance evaluation of Active Learner in the absence of ground truth.

A challenge within AL frameworks is the cold-start problem, which occurs due to absence of labels in the beginning for training an initial model. Among the alternatives, there is a proposal to incorporate uncertainty and diversity sampling into a unified process to select most representative samples as the initial labelled dataset \cite{rw01}. Gong et al. \cite{rw02} propose a novel inference method based on Bayesian Deep Latent Gaussian Model (BELGAM) to select initial training instances. Another option to address the cold-start problem is an unsupervised matching method that proposes bootstrapping active learning \cite{rw03}. Furthermore, Deng et al. \cite{rw04} introduced a sequence-based adversarial learning model to select initial set of training instances for the AL methods. In this paper, we propose an AL framework that employs a pre-clustering step to select the points closest to the centroids of each cluster as instances for initial labelled dataset.

The problem of oracle uncertainty occurs due to the assumption of oracle being infallible in AL methodologies. A majority of approaches proposed in the literature can be classified into two categories. The first category of AL methods \cite{rw05, rw06, rw07} contemplates a multi-annotator set-up and combines the confidence levels of multiple uncertain annotators to obtain a single confidence metric. The other category of approaches \cite{rw08, rw09, rw10} accounts for the oracle noise by adding denoising layers or including penalties in objective functions within the querying framework. In contrast to the prior efforts, we propose a generic framework to handle oracle uncertainty that builds upon the ambiguity surrounding the expertise of the labeler and the confidence in the given labels. The confidence scores associated with the model predictions, as well as the labels supplied by experts, are considered as a part of the querying process. Such a strategy to handle oracle uncertainty by incorporating additional feedback from the experts in the form of confidence levels has not been reported in the Active Learning literature so far to the best of our knowledge and stands out as a novel contribution of this work.

A majority of the AL methodologies report model performance in terms of classification accuracy, which requires ground truth information. However, it is not feasible to be informed about ground truth in practical scenarios. We propose the use of AL heuristics as surrogate metrics to evaluate the performance of the Active Learner. These metrics are observed to mimic the role of classification accuracy in model evaluation. The AL methodology presented in \cite{ALconfpaper01} addresses this issue along with handling oracle uncertainty. However, it still lacks the ability to solve cold-start problem in practical AL scenarios. The proposed AL framework possesses the capability to address multiple challenges simultaneously, i.e., cold-start, oracle uncertainty and performance evaluation of Active Learner in the absence of ground truth. Additionally, it supports the use of hybrid query strategies, rather than employing a single query strategy during the entire querying process. This work demonstrates that the use of hybrid query strategies helps in reducing the computational cost, while delivering at par or better performance than the individual query strategies. The values of AL heuristics are used to identify instances of switching between the query strategies. This approach of implementing hybrid query strategies has not been presented in the literature before and is an important contribution of this work. 

\section{Active Learning Challenges}

\subsection{Cold-start problem in Active Learning}

The initial labelled dataset is a key component in an Active Learning framework. The expectation is that a small portion of the dataset is labelled and available to the learner for its use. However, acquiring labels for the initial labelled dataset can be extremely time-consuming, expensive, or infeasible in many practical situations \cite{coldst1}. This gives rise to the cold-start problem in Active Learning settings, where there are no labels in the beginning to train an initial classification model. The proposed framework addresses this issue through a pre-clustering approach. In this approach, the unlabelled dataset is clustered using unsupervised clustering algorithm and the points closest to the centroids of each cluster are selected as the instances for an initial labelled dataset.

\subsection{Oracle Uncertainty}

The oracle is assumed to be infallible in Active Learning settings~\cite{oracletruth}. In practical scenarios, the labels are obtained from experimental test set-ups or provided by a human expert. In either case, there is an uncertainty associated with the labels due to experimental errors, measurement noise or human mistakes \cite{settles2011theories}. This motivates the issue of oracle uncertainty in conventional Active Learning methodologies. We address this issue by incorporating additional feedback during the query process. This feedback takes the form of (i) a grade on the expertise of the human oracle and (ii) a confidence level of the expert in providing such label. The feedback helps the Active Learner to quantify the reliability of annotations obtained from the oracle. 

\subsection{Hybrid query strategies}

Active Learning frameworks employ querying strategies to select instances. These strategies include Uncertainty Sampling (US) \cite{us01dwm01, tkdd_spec_02, tkdd_spec_03} and Query-by-Committee (QBC) \cite{qbc01, qbc02, kbs02_specif}. Other query strategies such as Expected Error Reduction \cite{eer01, tkdd_spec_04}, Variance Reduction \cite{vr01, vr02} and Density-weighted methods (DWM) \cite{us01dwm01, dwm02, tkdd_spec_05} are described in the literature. The selection of query strategy for a specific application depends on several factors such as computational cost, feasibility to implement and compatibility with base learning methods. Active Learning methodologies usually implement a single query strategy throughout the entire querying process. However, this is not convenient in practical scenarios. The reasons include a query strategy that might be easy to implement but fail to deliver satisfactory performance, or another strategy might exhibit stellar performance but possess high computational cost. This motivates the use of hybrid query strategies, which allows to combine the benefits of multiple methods within a single framework. For instance, a hybrid combination of two appropriate query strategies may help to reduce the computational cost, while exhibiting at par or better performance than the individual query strategies.    

\subsection{Absence of ground truth}

The performance evaluation of Active Learning methodologies is generally performed by comparing the predicted labels against the ground truth. Hence, the corresponding results are reported in the form of classification accuracy. However, acquiring ground truth labels for all the instances in a given dataset is not straightforward in practical settings due to the large amount of work involved in the labelling process. Moreover, the labelling process may also result in experimental or human errors, which degrade the quality of the labels. Due to the aforementioned shortcomings, classification accuracy is not an appropriate indicator of the performance of the Active Learner. This work proposes the use of heuristics evaluated during the iterative querying process, as metrics to mimic classification accuracy. These heuristics are observed to exhibit correlation with classification accuracy, and represent the performance of the Active Learner.

\section{Proposed Framework}

\begin{figure}
    \centering
    \includegraphics[width=\textwidth]{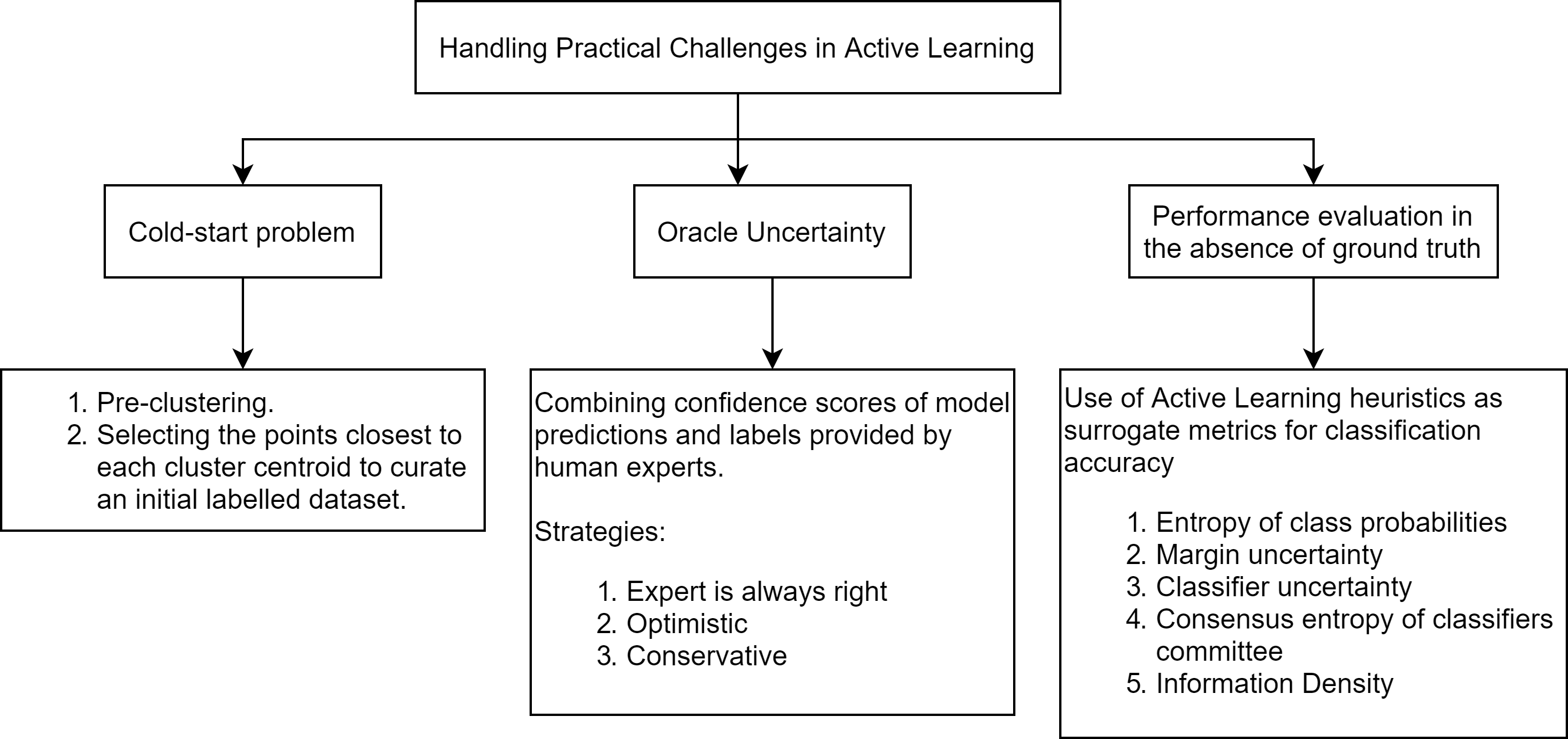}
    \caption{Proposed Active Learning Framework}
    \label{fig:fig1}
\end{figure}

\begin{figure}
    \centering
    \includegraphics[width=\textwidth]{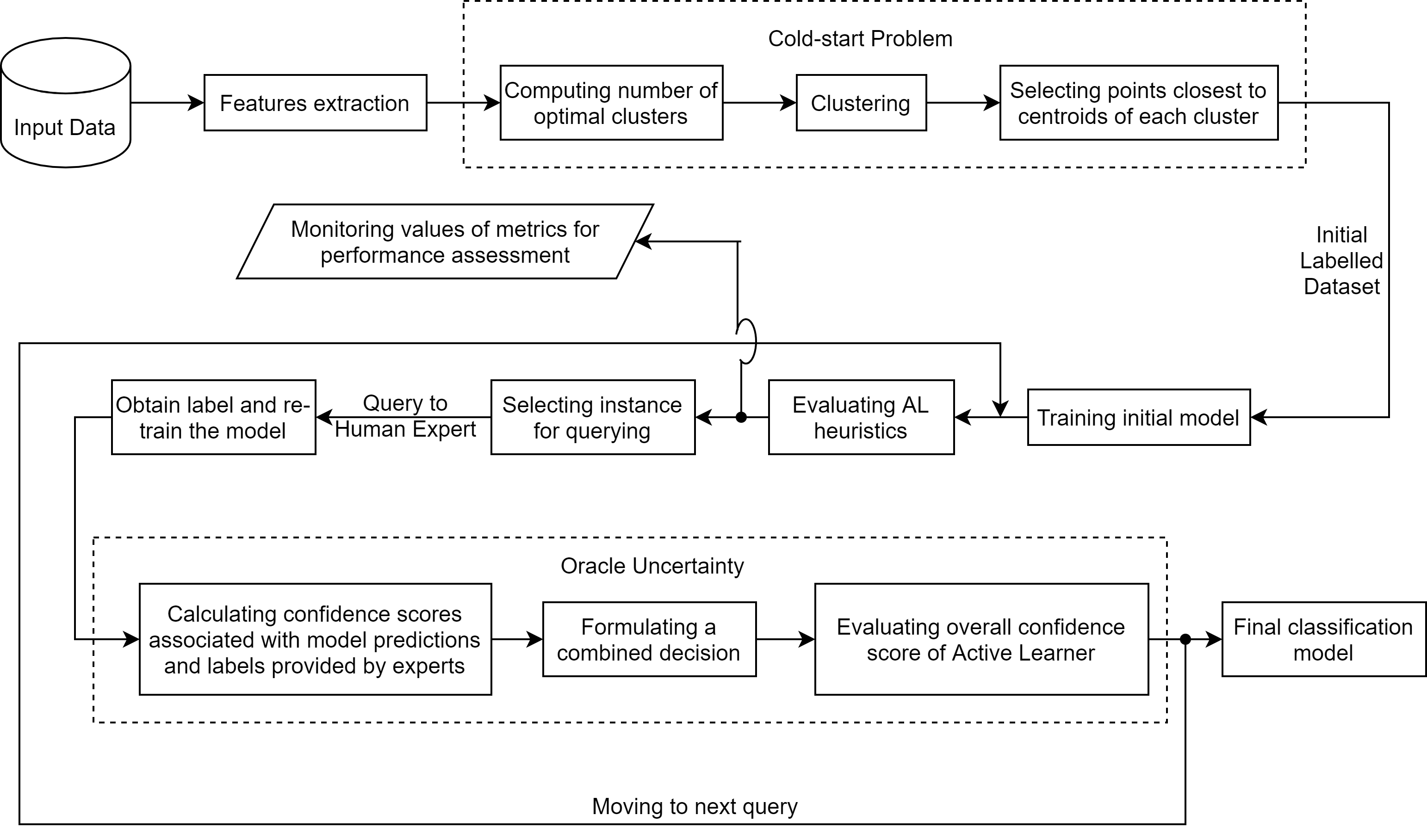}
    \caption{Flowchart for implementing the proposed AL framework}
    \label{fig:flowchart01}
\end{figure}

The proposed Active Learning framework incorporates a hybrid query strategy that addresses the following practical challenges: (i) cold-start problem, (ii) oracle uncertainty, (iii) use of hybrid query strategies, and (iv) performance quantification in the absence of ground truth labels. The proposed framework is presented in Figure \ref{fig:fig1}. We address the cold-start problem using pre-clustering to assign labels to instances close to the cluster centroids. Meanwhile, the issue of oracle uncertainty in handled by making a combined decision based on the confidence scores of model predictions and those of the human experts. In addition, we propose a set of metrics derived from Active Learning heuristics to evaluate the quality of the labels. These metrics are also used to identify the instance that requires switching to a different query algorithm under the proposed hybrid query strategy. A flowchart from implementing the proposed framework is demonstrated in Figure \ref{fig:flowchart01} and all the corresponding elements are elaborated as follows.     

\subsection{Addressing cold-start problem}
\label{meth:coldstart}

The proposed framework handles the cold-start problem in Active Learning by introducing a pre-clustering step that generates the initial labelled dataset. Given a dataset with no labels, the procedure is as follows:

\begin{enumerate}
    \item Select an unsupervised clustering algorithm and apply it to the dataset with no labels. In this work, k-means clustering has been used for the pre-clustering step as it is relatively simple to implement and scales to large datasets easily.
    \item Identify the number of optimal clusters for the given dataset. The elbow method is employed to select the optimal number of clusters in this work, as discussed in \cite{tkdd_spec_06}.
    \item Create the initial labelled dataset by selecting the points closest to the centroids of each cluster, such that the number of instances in the generated set do not exceed a pre-defined fraction of that in the complete dataset. This work considers the threshold to be 2\%, i.e., the size of the initial labelled dataset should be 2\% of the entire dataset.
\end{enumerate}

Once the initial labelled dataset is obtained using the aforementioned procedure, it can be used to train the classification model, which is updated continuously during the iterative querying process in the Active Learning framework.

\subsection{Handling oracle uncertainty}
\label{meth:oracleuncert}

The oracle uncertainty challenge originates in the ambiguity surrounding the expertise of the labeler and the confidence in the given labels. Thus, a solution to this challenge requires a way to incorporate this uncertainty via a confidence score that accounts for the model predictions and the input labels. 
The Active Learning process produces a posterior probability associated to each class for all the instances. The posterior probabilities indicate the likelihood of a particular instance being classified into the given class per the current model. The posterior probability can be used to calculate heuristics such as the level of uncertainty, entropy of class probabilities and classification margin. The maximum value of the posterior probability is calculated using

\begin{equation}
    p_{max} = \max \left( P_\theta (\hat{y} | \bm{x}) \right),
    \label{eq:eq1}
\end{equation}

\noindent where $\hat{y}$ represents the predicted label for the instance $\bm{x}$ under the model $\theta$. Further, $p_{max} \in [0, 1]$ is scaled-up to the range of 1-5 to associate confidence scores with the model predictions. Here, 1 represents the lowest confidence score and 5 represents the
highest confidence score.

In addition to the confidence scores of the model, the confidence scores supplied by the oracle need to be incorporated as part of this framework. The confidence score associated with labels supplied by the human experts is a result of the following inputs during the continuous querying process: (i) label, (ii) grade (level of expertise) of the human annotator ($z_1$) and (iii) confidence level corresponding to the label provided ($z_2$). $z_1$ and $z_2$ are numeric values between 1-5, where one represents the lowest expertise/confidence level and five represents the highest expertise/confidence level. Variables $z_1$ and $z_2$ are used to produce an overall confidence score of the expert on a specific label, as per the rule base shown in Table \ref{tab:ruleBase}. It should be noted that the lower graded experts are penalized by imposing a lower confidence score than what is originally reported ($z_2$). This is done because we have relatively lesser trust on the lower graded experts. On the other hand, the scores provided by higher graded experts are accepted without any penalization.  

\begin{table}
\centering
\caption{Rule to generate overall confidence score of the expert}
\label{tab:ruleBase}
\resizebox{0.4\columnwidth}{!}{%
\begin{tabular}{|c|c|c|c|c|c|} 
\hline
\multirow{2}{*}{\begin{tabular}[c]{@{}c@{}}\textbf{Confidence}\\\textbf{Level ($z_2$)} \end{tabular}} & \multicolumn{5}{c|}{\textbf{Grade of Expert ($z_1$)}}           \\ 
\cline{2-6}
                                                                                                      & \textbf{1} & \textbf{2} & \textbf{3} & \textbf{4} & \textbf{5}  \\ 
\hline
\textbf{Not Provided}                                                                                 & 1          & 1          & 2          & 2          & 3           \\ 
\hline
\textbf{1}                                                                                            & 1          & 1          & 1          & 1          & 1           \\ 
\hline
\textbf{2}                                                                                            & 1          & 1          & 1          & 2          & 2           \\ 
\hline
\textbf{3}                                                                                            & 1          & 1          & 2          & 3          & 3           \\ 
\hline
\textbf{4}                                                                                            & 1          & 2          & 2          & 4          & 4           \\ 
\hline
\textbf{5}                                                                                            & 1          & 2          & 3          & 4          & 5           \\
\hline
\end{tabular}
}
\end{table}

The next step involves formulation of a combined decision based on the confidence scores of model predictions and the labels provided by human experts. This is implemented using three different strategies discussed as follows:

\begin{enumerate}
    \item \textbf{Expert is always right:} The confidence score corresponding to the label provided by the expert is given a higher priority and is used to update the confidence score of the model during each query.
    \item \textbf{Optimistic approach:} It considers the higher confidence score among that of the model and the expert.
    \item \textbf{Conservative approach:} It considers the lower confidence score among that of the model and the expert.
\end{enumerate}

Finally, the confidence scores of model and the expert are used to generate an overall confidence score of the Active Learner at each query step. It is calculated as the weighted average of the scores corresponding to all the instances in the dataset, as given by

\begin{equation}
    S_{AL} = \frac{\sum_{i = 1}^5 s_i n_i}{\sum_{i = 1}^5 n_i},
    \label{eq:eq2}
\end{equation}

\noindent where, $n_i$ represents the number of instances in the dataset with confidence score $s_i$.

\subsection{Performance evaluation and hybrid query strategies}
\label{meth:perfhybrid}

The fundamental premise of Active Learning methodologies is the iterative selection of most informative instances by the optimization of selected heuristics. The framework here described proposes the use of the selected heuristics as metrics to quantify the performance of the Active Learner. This approach is beneficial in many practical scenarios, where it is not feasible to obtain ground truth labels for all the instances in a dataset. The metrics used in this work are presented in Table~\ref{tab:metricsDetails}. Entropy of the class probabilities (EC), Margin uncertainty (MU) and Classifier uncertainty (CU) are associated with uncertainty measures in US query strategy. Consensus entropy of the committee of classifiers (CE) is derived from disagreement measures in QBC. Information densities with Euclidean distance (IE) and cosine similarity (IC) are linked to the similarity measures in DWM approach. 

Given that these metrics are derived from the heuristics involved in Active Learning settings, they are expected to mimic classification accuracy so that the performance of Active Learner can be evaluated in the absence of ground truth. For instance, the value of classifier uncertainty decreases with the most informative instances being labelled at each query step. Consequently, the value of CU should decrease with an increase in classification accuracy as more instances are labelled iteratively. Similarly, the values of EC and CE are also expected to decrease and MU is expected to increase in relation to classification accuracy as more queries are made. The performance evaluation of Active Learning methodologies using such heuristics has not been reported in the relevant literature so far to the best of our knowledge and stands out as a novel contribution of this work.

In this work, the metrics CU, MU, EC and CE are used, which can be calculated as follows:
\begin{itemize}
    \item CU can be calculated by
    \begin{equation}
        \text{CU} = 1 - P_\theta(\hat{y} | \bm{x})
    \end{equation}
    \item MU can be written as
    \begin{equation}
        \text{MU} = P_\theta(\hat{y_1} | \bm{x}) - P_\theta(\hat{y_2} | \bm{x})
    \end{equation}
    where, $\hat{y_1}$ and $\hat{y_2}$ are the first and second most likely predictions under the model $\theta$, respectively.
    \item EC can be evaluated as
    \begin{equation}
        \text{EC} = -\sum_y P_\theta(y | \bm{x}) \log(P_\theta(y | \bm{x}))
    \end{equation}
    where $y$ ranges over all possible labelling of $x$.
    \item CE can be expressed as
    \begin{equation}
        \text{CE} = -\sum_y P_\mathcal{C}(y | \bm{x}) \log(P_\mathcal{C}(y | \bm{x}))
    \end{equation}
    where,
    \begin{equation}
        P_\mathcal{C}(y | \bm{x}) = \frac{1}{|\mathcal{C}|} \sum_{\theta \in \mathcal{C}} P_\theta(y | \bm{x})
    \end{equation}
    $\mathcal{C}$ is the committee of classifiers and $|\mathcal{C}|$ is the size of the committee.
    \item IE can be computed as
    \begin{equation}
        \text{IE} = \frac{1}{|\mathcal{U}|} \sum_{\bm{x'} \in \mathcal{U}} d(\bm{x}, \bm{x'})
    \end{equation}
    where $\mathcal{U}$ represents the pool of unlabelled instances, $|\mathcal{U}|$ is the size of the pool of unlabelled instances and $d(\bm{x}, \bm{x'})$ is the Euclidean distance between $\bm{x}$ and $\bm{x'}$, given by
    \begin{equation}
        d(\bm{x}, \bm{x'}) = \sqrt{(x_1 - x'_1)^2 + (x_2 - x'_2)^2 + ... + (x_n - x'_n)^2}
    \end{equation}
    $n$ is the size of feature vectors used for training the classification models.
    \item IC can be determined as
    \begin{equation}
        \text{IC} = \frac{1}{|\mathcal{U}|} \sum_{\bm{x'} \in \mathcal{U}} sim(\bm{x}, \bm{x'})
    \end{equation}
    where $sim(\bm{x}, \bm{x'})$ is the cosine similarity between $\bm{x}$ and $\bm{x'}$, given by
    \begin{equation}
        sim(\bm{x}, \bm{x'}) = \frac{\bm{x} \cdot \bm{x'}}{||\bm{x}|| \, ||\bm{x'}||}
    \end{equation}
\end{itemize}
 
 The values of the metrics enlisted in Table \ref{tab:metricsDetails} can also be examined to identify suitable instances of switching to a different query strategy rather than using a same strategy throughout the querying process. This approach is highly advantageous in numerous industrial settings, where it is desirable to have a trade-off between aspects such as labelling expenditures, computational cost and appreciable performance of the Active Learner. The query strategy can be switched when the values of metrics either oscillate beyond pre-defined thresholds or do not exhibit a substantial change over a certain number of queries. Moreover, the decision regarding termination of querying process can also be made when the metrics values exhibit saturation or approach a pre-determined threshold. The use of hybrid strategies is motivated by the equal or improved performance when compared to individual query strategies at a lower computational cost.  

\begin{table}[]
\caption{Details of the proposed metrics}
\label{tab:metricsDetails}
\begin{tabular}{|c|l|c|l|}
\hline
\textbf{Serial Number} & \multicolumn{1}{c|}{\textbf{Metric Description}} & \textbf{Abbreviation} & \multicolumn{1}{c|}{\textbf{Remarks}} \\ \hline
1 & Entropy of the class probabilities & EC & \multirow{3}{*}{\begin{tabular}[c]{@{}l@{}}Measures of \\ uncertainty\end{tabular}} \\ \cline{1-3}
2 & \begin{tabular}[c]{@{}l@{}}Margin uncertainty, i.e., difference \\ of the probabilities of first and second \\ most likely predictions\end{tabular} & MU &  \\ \cline{1-3}
3 & \begin{tabular}[c]{@{}l@{}}Classifier uncertainty, i.e., \\ 1 - P(correct prediction)\end{tabular} & CU &  \\ \hline
4 & \begin{tabular}[c]{@{}l@{}}Consensus entropy of the committee \\ of classifiers in QBC approach\end{tabular} & CE & \begin{tabular}[c]{@{}l@{}}Measure of \\ disagreement\end{tabular} \\ \hline
5 & \begin{tabular}[c]{@{}l@{}}Information density with Euclidean \\ distance as similarity measure\end{tabular} & IE & \multirow{2}{*}{\begin{tabular}[c]{@{}l@{}}Density-\\ weighted \\ heuristics\end{tabular}} \\ \cline{1-3}
6 & \begin{tabular}[c]{@{}l@{}}Information density with cosine \\ similarity as similarity measure\end{tabular} & IC &  \\ \hline
\end{tabular}
\end{table}

\section{Experimental Evaluation}

We are interested to evaluate the robustness of our proposed Active Learning framework across different environments and industrial settings. Therefore, we implement the framework in the following datasets: (i) process data collected from a simulated ethanol plant \cite{cstrDataRef}, (ii) drug consumption (quantified) dataset from UCI Machine Learning repository \cite{dataset02}, and (iii) rolling element bearing test data from the bearing data center at Case Western Reserve University \cite{dataset01}. The details of the analyses and results are described below.

\subsection{Description of the datasets}

\subsubsection{Dataset 1 - Process data collected from a simulated ethanol plant:}

We use data from a simulated ethanol plant to describe how to overcome the clod-start challenge. The simulations are based on mass and energy balances and the corresponding Human Machine Interface (HMI) is designed using MATLAB. Details of the simulated ethanol plant are presented in \cite{cstrDataRef}. It consists of a continuous stirred tank reactor (CSTR) and a distillation column (DC). The measurements available correspond to 11 process variables: ethanol concentration in CSTR (C101), feed flow rate to CSTR (F101) and DT (F105), coolant flow rate (F102), level of CSTR (L101), temperatures of coolant inlet (T101), coolant outlet (T102), CSTR (T103), third tray of DT (T104), fifth tray of DT (T105) and eighth tray of DT (T106) \cite{cstrDataRef}. There are 4 different scenarios within these measurements: normal operating conditions (14041 instances), coolant-related disturbances (1647), CSTR-related disturbances (2819) and reflux imbalance (9336), thereby constituting a total of 27843 instances. The output of these 11 processes are treated as features to the classification problem. The output data originates from an experiment where 10 participants were asked to control the process variables manually if their values extend beyond the normal operating range.


\subsubsection{Dataset 2 - Drug consumption (quantified) dataset:}

This dataset contains details of 1885 participants regarding their usage of various types of legal and illegal drugs. The features for this dataset comprise of personality measurements, which are termed as NEO-FFI-R (neuroticism, extraversion, openness to experience, agreeableness, and conscientiousness), BIS-11 (impulsivity), and ImpSS (sensation seeking). In addition, it includes level of education, age, gender, country of residence and ethnicity as features too \cite{dataset02}. The categorical input attributes are quantified and used as numerical values \cite{dataset02}. The labels contain information about usage of cannabis drugs and consist of seven classes: "Never Used", "Used over a Decade Ago", "Used in Last Decade", "Used in Last Year", "Used in Last Month", "Used in Last Week", and "Used in Last Day". The classes are reorganized to form a "year-based" binary classification problem as described in \cite{dataset02}: the classes "Never Used", "Used over a Decade Ago" and "Used in Last Decade" are merged to form a group of cannabis non-users and all other classes are combined to form a group of cannabis users. 

\subsubsection{Dataset 3 - Rolling element bearing test data:}

This dataset consists of ball bearing test data obtained from experiments conducted using a 2 hp electric motor under normal and faulty conditions. The faults vary in size (0.007" to 0.040" in diameter) as well as location (inner raceway, outer raceway, or rolling element). We use the vibration data of the motor captured at the drive end for loads in the range 0-3 hp. This corresponds to four different motor speeds between 1720-1797 RPM. The data is sampled at 12 kHz for 10 seconds at each speed level. In this experiment, each of these signal segments are split into 40 instances of 0.25 seconds each at each speed level. It results in 160 (40 x 4 speed levels) instances for each normal and fault condition. Furthermore, the classification problem is formulated by selecting the fault size as 0.021” and having four classes corresponding to the fault location - healthy, inner-race, ball and outer-race. Consequently, there are a total of 640 (160 x 4 classes) instances used in this experiment. Wavelet Packet Decomposition (WPD) technique is employed for extracting features from these signals, as discussed in \cite{wpd}. WPD is performed at level 6 using mother wavelet from Daubechies family as it has been reported to be used in other health management problems \cite{wpd}. Furthermore, the wavelet coefficients that correspond to different frequency bands are extracted and their Shannon entropy values are used as features for classification problem. WPD analysis is performed using the Wavelet toolbox in MATLAB.

\subsection{Experimental Settings}

\subsubsection{Dataset 1 - Process data collected from a simulated ethanol plant:}

\begin{figure}
    \centering
    \includegraphics[width=0.6\textwidth]{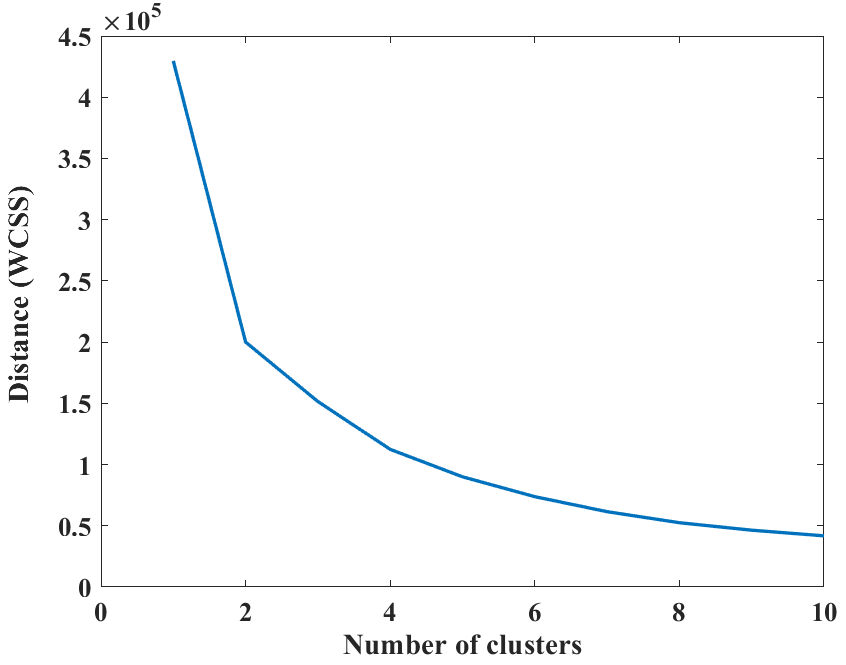}
    \caption{Plot of WCSS vs. number of clusters for Dataset 2: Drug consumption (quantified) dataset.}
    \label{fig:fig2}
\end{figure}

The initial labelled dataset is pre-processed with the method described in Section \ref{meth:coldstart}. The optimal number of clusters is identified as four using the elbow method. The plot of within-cluster sum of squares distance (WCSS) vs. number of clusters is shown in Figure \ref{fig:fig2}. It can be observed that after four clusters, the rate of change in the WCSS distance decreases significantly. An “elbow” is visible at four-clusters point. The size of the initial labelled dataset is set as 500 instances with 125 instances for each class.

Once the initial labelled dataset is formed, the first model is trained and the iterative querying process begins. QBC is implemented in Python using the modAL framework \cite{modal} as the basic tool for selection of instances for annotation by a human expert. kNN is selected as the base classification method because it is simple and easy to implement, versatile, and is a non-parametric algorithm. Moreover, it does not make any assumptions on the input data distribution. The metrics introduced in Section \ref{meth:perfhybrid} are also evaluated at each query step along with the classification accuracy.

\subsubsection{Dataset 2 - Drug consumption (quantified) dataset:}

This is a binary classification problem, where the given participant may be a cannabis user or non-user. The initial labelled dataset is created by selecting 2\% of the total instances at random along with corresponding labels. An initial model is formed by using the initial labelled dataset and then the iterative querying process is initiated. In these experiments, US and QBC are implemented for querying in Python using the modAL framework \cite{modal} and kNN is chosen as the base classifier. In the experiment using this dataset, three query procedures are used. These procedures are: (i) all queries using US, (ii) all queries using QBC, and (iii) hybrid approach, i.e., first half of the queries using US and another half using QBC. All the relevant metrics are recorded and monitored at each query step for performance evaluation in the absence of ground truth and selection of instances for switching between the query strategies.

\subsubsection{Dataset 3 - Rolling element bearing test data:}

The dataset is pre-processed using WPD to obtain the features for all the instances. The procedure described in Section \ref{meth:coldstart} is applied to the resulting instances to overcome the cold-start problem. Using the elbow method, the optimal number of clusters is set to four. Once the initial labelled dataset is available, the first model is trained and the iterative querying process is initiated. The modAL framework \cite{modal} is used for implementation of US and QBC query strategies.

A total of 640 normally distributed random numbers in the range of 1-5 are used to generate variables $z_1$ and $z_2$. The normally distributed random numbers help to simulate the difficulty of finding many maintenance and vibration experts with different levels of skills. The resulting overall confidence score of the expert is produced using variables $z_1$ and $z_2$ according to the procedure described in Section \ref{meth:oracleuncert}. Afterwards, the overall confidence score of the Active Learner is calculated using Eq. \ref{eq:eq2} for all the three strategies enlisted in Section \ref{meth:oracleuncert}. Finally, the relevant metrics are calculated and monitored at each query step for different purposes, as highlighted in Section \ref{meth:perfhybrid}.

\subsection{Results}

The advantages of implementing a pre-clustering approach to alleviate the cold-start problem is illustrated using experiments performed on ethanol plant dataset. Table \ref{tab:resD2} presents the results of the Active Learning framework within the context of the cold-start challenge with the pre-clustering step incorporated. The entries in the column \textit{Unqueried value} represent the metrics values when the model is trained with only the initial labelled dataset and before any queries were carried out. The entries in subsequent columns indicate the metrics values corresponding to the updated model after reaching the indicated number of queries. Firstly, it can be observed that the classification accuracy increases as more queries are made. This indicates that the performance of the Active Learner improves during the iterative querying process. Moreover, it is worthwhile to note that classification accuracy reaches 91\% with 2000 queries, which is just around 7\% of the total instances in the dataset.

\begin{table}
\centering
\caption{Results for Dataset 1: Process data collected from a simulated ethanol plant}
\label{tab:resD2}
\resizebox{0.7\columnwidth}{!}{%
\begin{tabular}{|c|c|c|c|c|c|c|} 
\hhline{|=======|}
\multirow{2}{*}{\textbf{Metrics}}                                                                                                                             & \multirow{2}{*}{\begin{tabular}[c]{@{}c@{}}\textbf{Unqueried }\\\textbf{Value}\end{tabular}} & \multicolumn{5}{c|}{\textbf{Values after making no. of queries}}              \\ 
\cline{3-7}
                                                                                                                                                              &                                                                                              & \textbf{500} & \textbf{1000} & \textbf{1500} & \textbf{2000} & \textbf{2500}  \\ 
\hhline{|=======|}
\textbf{EC}                                                                                                                                             & 0.38                                                                                         & 0.35         & 0.29          & 0.26          & 0.23          & 0.21           \\ 
\hline
\textbf{MU}                                                                                                                                             & 0.81                                                                                         & 0.83         & 0.84          & 0.86          & 0.87          & 0.9            \\ 
\hline
\textbf{CU}                                                                                                                                             & 0.14                                                                                         & 0.12         & 0.11          & 0.1           & 0.08          & 0.07           \\ 
\hline
\textbf{CE}                                                                                                                                             & 0.18                                                                                         & 0.15         & 0.11          & 0.09          & 0.06          & 0.04           \\ 
\hline
\textbf{IE}                                                                                                                                             & 0.05                                                                                         & 0.05         & 0.06          & 0.06          & 0.06          & 0.07           \\ 
\hline
\textbf{IC}                                                                                                                                             & 0.96                                                                                         & 0.97         & 0.97          & 0.97          & 0.98          & 0.98           \\ 
\hhline{|=======|}
\begin{tabular}[c]{@{}c@{}}\textbf{Classification }\\\textbf{Accuracy}\end{tabular}                                                                           & 0.71                                                                                         & 0.77         & 0.81          & 0.87          & 0.91          & 0.97           \\ 
\hline
\begin{tabular}[c]{@{}c@{}}\textbf{Classification }\\\textbf{Accuracy - using }\\\textbf{randomly generated }\\\textbf{initial labelled dataset}\end{tabular} & 0.69                                                                                         & 0.72         & 0.77          & 0.84          & 0.89          & 0.96           \\
\hhline{|=======|}
\end{tabular}
}
\end{table}

Table \ref{tab:resD2} shows that EC, MU, CU and CE exhibit a correlation with the classification accuracy. EC, CU and CE are observed to decrease with increased classification accuracy as more instances are queried. This is expected because entropy of the class probabilities, level of uncertainty and level of disagreement between the committee members (QBC approach) decrease as the most informative instances are labelled by the Active Learner during the initial set of iterations. On the other hand, MU grows with increased classification accuracy as more instances are queried. The reason is a higher margin implies less uncertainty in distinguishing between the two most likely alternatives and is likely to increase with increasing number of queries. The metrics IE and IC do not exhibit an appreciable change as more instances are queried because information density is not considered while selecting instances using QBC query strategy. These heuristics might change when using Density Weighted Methods for instance selection during querying. Hence, the proposed framework can be used to evaluate the performance of Active Learner using the metrics in the absence of ground truth information. Secondly, the classification accuracies corresponding to the proposed framework is a bit higher as compared to the case when the initial labelled dataset would have been obtained by means of random sampling. This is an added advantage of the pre-clustering step in the proposed Active Learning framework.

The benefits of hybrid query strategies are established using experiments performed on the drug consumption dataset and the corresponding results are shown in Table \ref{tab:hybridqsD1}. Here, 500 queries (i.e., around 25\% of total instances in the dataset) are made using three different modes: (i) all queries using US, (ii) all queries using QBC, and (iii) a hybrid of US and QBC - first half of the queries using US and next half using QBC. In all the querying procedures, the classification accuracy increases as more queries are made, thereby indicating the improving performance of the Active Learner during the querying process.  

\begin{table}[t]
\centering
\caption{Results for hybrid query strategy - Dataset 2: Drug consumption (quantified) dataset}
\label{tab:hybridqsD1}
\resizebox{\columnwidth}{!}{%
\begin{tabular}{|c|c|c|c|c|c|c|c|c|c|c|c|c|} 
\hhline{|=============|}
\multirow{3}{*}{\textbf{Metrics}}                                                   & \multicolumn{2}{c|}{\multirow{2}{*}{\begin{tabular}[c]{@{}c@{}}\textbf{Unqueried }\\\textbf{value}\end{tabular}}} & \multicolumn{10}{c|}{\textbf{Values after making no. of queries}}                                                                                                                          \\ 
\cline{4-13}
                                                                                    & \multicolumn{2}{c|}{}                                                                                             & \multicolumn{2}{c|}{\textbf{125}} & \multicolumn{2}{c|}{\textbf{250}} & \multicolumn{3}{c|}{\textbf{375}}                       & \multicolumn{3}{c|}{\textbf{500}}                        \\ 
\cline{2-13}
                                                                                    & \textbf{US} & \textbf{QBC}                                                                                        & \textbf{US} & \textbf{QBC}        & \textbf{US} & \textbf{QBC}        & \textbf{US} & \textbf{QBC} & \textbf{\textit{US + QBC}} & \textbf{US} & \textbf{QBC} & \textbf{\textit{US + QBC}}  \\ 
\hhline{|=============|}
\textbf{EC}                                                                   & 0.64        & 0.58                                                                                                & 0.56        & 0.51                & 0.52        & 0.48                & 0.47        & 0.44         & \textit{0.46}              & 0.43        & 0.41         & \textit{0.41}               \\ 
\hline
\textbf{MU}                                                                   & 0.27        & 0.38                                                                                                & 0.34        & 0.44                & 0.42        & 0.49                & 0.48        & 0.56         & \textit{0.5}               & 0.52        & 0.63         & \textit{0.57}               \\ 
\hline
\textbf{CU}                                                                   & 0.36        & 0.32                                                                                                & 0.33        & 0.29                & 0.31        & 0.24                & 0.27        & 0.21         & \textit{0.26}              & 0.25        & 0.19         & \textit{0.23}               \\ 
\hline
\textbf{CE}                                                                   & 0.53        & 0.48                                                                                                & 0.46        & 0.42                & 0.43        & 0.39                & 0.39        & 0.36         & \textit{0.38}              & 0.35        & 0.34         & \textit{0.34}               \\ 
\hline
\textbf{IE}                                                                   & 0.21        & 0.21                                                                                                & 0.21        & 0.22                & 0.22        & 0.22                & 0.22        & 0.23         & \textit{0.23}              & 0.23        & 0.23         & \textit{0.23}               \\ 
\hline
\textbf{IC}                                                                   & 0.52        & 0.51                                                                                                & 0.52        & 0.51                & 0.52        & 0.51                & 0.51        & 0.51         & \textit{0.51}              & 0.51        & 0.5          & \textit{0.51}               \\ 
\hhline{|=============|}
\begin{tabular}[c]{@{}c@{}}\textbf{Classification }\\\textbf{Accuracy}\end{tabular} & 0.58        & 0.69                                                                                                & 0.64        & 0.74                & 0.76        & 0.8                 & 0.85        & 0.88         & \textit{0.89}              & 0.89        & 0.9          & \textit{0.91}               \\
\hhline{|=============|}
\end{tabular}
}
\end{table}

It can be seen from the values of classification accuracy that the performance of the Active Learner in the hybrid case is at par or better than that of the individual query strategies. The US strategy has greater ease of implementation when compared to QBC but it provides lower performance results than QBC. On the other hand, QBC is computationally expensive when compared to US, although it achieves a superior performance in terms of classification accuracy. The hybrid approach of US + QBC helps to accomplish a trade-off between these two aspects - the performance at the end of 500 queries is better than that of individual query strategies, while simultaneously reducing the expense of QBC as first 250 queries are made using US, which is comparatively cheaper in terms of computational cost. Finally, all the metrics for the three querying procedures exhibit a correlation with classification accuracy as shown in Table~\ref{tab:hybridqsD1}. This correlation is an expected behavior given the relationship between the metrics and classification accuracy.

The contribution of all the elements together in the proposed framework is exhibited through experiments performed on rolling element bearing dataset. The results of this experiment that aims to evaluate the performance of an hybrid query strategy are shown in Table \ref{tab:hybridqsD3}, where the hybrid combination of US + QBC is observed to deliver at par or better performance than individual query strategies. The trade-off between model quality and computational cost provides an additional incentive for the use of the hybrid strategy. When starting with a dataset without labels, the classification accuracy, relevant metrics and the overall confidence scores of the Active Learner using three combination strategies are reported in Table \ref{tab:resD3}. EC, CU and CE exhibit an inversely proportional correlation with classification accuracy and MU is directly proportional to classification accuracy as more queries are made. This behavior of the metrics is consistent with the previous dataset experiments.

\begin{figure}
    \centering
    \includegraphics[width=0.9\textwidth]{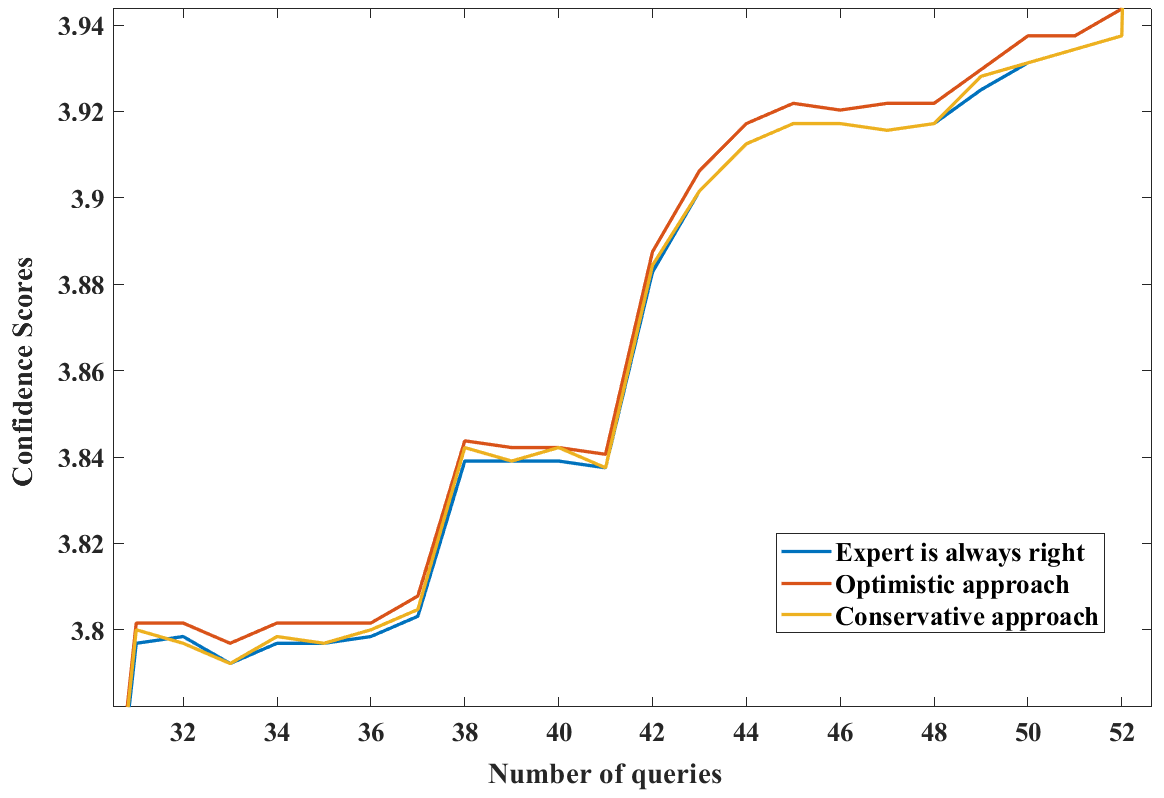}
    \caption{Plot of overall model confidence scores with number of queries made to the expert~- Dataset 3: Rolling element bearing test data from the bearing data center at Case Western Reserve University.}
    \label{fig:plotCase1}
\end{figure}

Furthermore, it can be observed that the overall confidence scores of the Active Learner obtained by combining those of the model predictions and the experts exhibit an increasing trend with increasing number of queries for all the strategies. The plot showing this performance over 30-52 queries is presented in Figure \ref{fig:plotCase1}. This range of queries is selected for better readability and comparing performance across different combination strategies. The intermediate valleys in the plot are probably due to outliers being selected by the Active Learner for annotation by the human experts. It can be observed that the overall confidence scores corresponding to the optimistic strategy is generally higher than the other strategies. This is intuitive as the optimistic approach considers higher confidence scores among that of the model and the expert.

\begin{table}
\centering
\caption{Results for hybrid query strategy - Dataset 3: Rolling element bearing test data from the bearing data center at Case Western Reserve University}
\label{tab:hybridqsD3}
\resizebox{0.95\columnwidth}{!}{%
\begin{tabular}{|c|c|c|c|c|c|c|c|c|c|c|c|c|} 
\hhline{|=============|}
\multirow{3}{*}{\textbf{Metrics }}                                                 & \multicolumn{2}{c|}{\multirow{2}{*}{\begin{tabular}[c]{@{}c@{}}\textbf{Unqueried }\\\textbf{value}\end{tabular}}} & \multicolumn{10}{c|}{\textbf{Values after making number of queries}}                                                                                                                     \\ 
\cline{4-13}
                                                                                   & \multicolumn{2}{c|}{}                                                                                             & \multicolumn{2}{c|}{\textbf{25}} & \multicolumn{2}{c|}{\textbf{50}} & \multicolumn{3}{c|}{\textbf{75}}                        & \multicolumn{3}{c|}{\textbf{100}}                        \\ 
\cline{2-13}
                                                                                   & \textbf{US} & \textbf{QBC}                                                                                        & \textbf{US} & \textbf{QBC}       & \textbf{US} & \textbf{QBC}       & \textbf{US} & \textbf{QBC} & \textbf{\textit{US + QBC}} & \textbf{US} & \textbf{QBC} & \textbf{\textit{US + QBC}}  \\ 
\hhline{|=============|}
\textbf{EC}                                                                  & 0.93        & 0.72                                                                                                & 0.77        & 0.64               & 0.62        & 0.52               & 0.53        & 0.44         & \textit{0.47}              & 0.48        & 0.38         & \textit{0.34}               \\ 
\hline
\textbf{MU}                                                                  & 0.15        & 0.32                                                                                                & 0.28        & 0.39               & 0.39        & 0.51               & 0.49        & 0.59         & \textit{0.51}              & 0.53        & 0.65         & \textit{0.59}               \\ 
\hline
\textbf{CU}                                                                  & 0.51        & 0.37                                                                                                & 0.45        & 0.32               & 0.41        & 0.25               & 0.37        & 0.23         & \textit{0.28}              & 0.27        & 0.19         & \textit{0.18}               \\ 
\hline
\textbf{CE}                                                                  & 0.89        & 0.67                                                                                                & 0.73        & 0.59               & 0.61        & 0.51               & 0.50        & 0.43         & \textit{0.43}              & 0.46        & 0.34         & \textit{0.30}               \\ 
\hline
\textbf{IE}                                                                  & 0.01        & 0.01                                                                                                & 0.02        & 0.02               & 0.02        & 0.01               & 0.02        & 0.02         & \textit{0.02}              & 0.03        & 0.02         & \textit{0.03}               \\ 
\hline
\textbf{IC}                                                                  & 0.58        & 0.58                                                                                                & 0.58        & 0.58               & 0.59        & 0.58               & 0.59        & 0.59         & \textit{0.59}              & 0.60        & 0.59         & \textit{0.60}               \\ 
\hhline{|=============|}
\begin{tabular}[c]{@{}c@{}}\textbf{Classification}\\\textbf{Accuracy}\end{tabular} & 0.40        & 0.53                                                                                                & 0.54        & 0.67               & 0.69        & 0.86               & 0.74        & 0.93         & \textit{0.89}              & 0.88        & 0.98         & \textit{0.99}               \\
\hhline{|=============|}
\end{tabular}
}
\end{table}

\begin{table}
\centering
\caption{Results for Dataset 3: Rolling element bearing test data from the bearing data center at Case Western Reserve University}
\label{tab:resD3}
\resizebox{0.7\columnwidth}{!}{%
\begin{tabular}{|c|c|c|c|c|c|c|} 
\hhline{|=======|}
\multirow{2}{*}{ \textbf{Metrics} }                                                                                                                        & \multirow{2}{*}{\begin{tabular}[c]{@{}c@{}}\textbf{Unqueried}\\\textbf{Value} \end{tabular}} & \multicolumn{5}{c|}{\textbf{Values after making no. of queries} }                       \\ 
\cline{3-7}
                                                                                                                                                           &                                                                                              & \textbf{25}  & \textbf{50}  & \textbf{75}  & \textbf{100}              & \textbf{125}   \\ 
\hhline{|=======|}
\begin{tabular}[c]{@{}c@{}}\textbf{Confidence Score}\\\textbf{(Expert is always right)} \end{tabular}                                                                  & 3.02                                                                                         & 3.6          & 3.93         & 4.45         & \multicolumn{1}{l|}{4.46} & 4.70           \\ 
\hline
\begin{tabular}[c]{@{}c@{}}\textbf{Confidence Score}\\\textbf{(Optimistic approach)} \end{tabular}                                                                  & 3.03                                                                                         & 3.64         & 3.94         & 4.46         & 4.46                      & 4.71           \\ 
\hline
\begin{tabular}[c]{@{}c@{}}\textbf{Confidence Score}\\\textbf{(Conservative approach)} \end{tabular}                                                                  & 3.03                                                                                         & 3.64         & 3.93         & 4.46         & 4.46                      & 4.70           \\ 
\hhline{|=======|}
\textbf{EC}                                                                                                                                          & 0.93                                                                                         & 0.80         & 0.66         & 0.57         & 0.42                      & 0.38           \\ 
\hline
\textbf{MU}                                                                                                                                          & 0.16                                                                                         & 0.24         & 0.40         & 0.57         & 0.75                      & 0.78           \\ 
\hline
\textbf{CU}                                                                                                                                          & 0.47                                                                                         & 0.41         & 0.38         & 0.31         & 0.25                      & 0.16           \\ 
\hline
\textbf{CE}                                                                                                                                          & 0.84                                                                                         & 0.74         & 0.61         & 0.51         & 0.41                      & 0.38           \\ 
\hline
\textbf{IE}                                                                                                                                          & 0.01                                                                                         & 0.01         & 0.02         & 0.02         & 0.02                      & 0.03           \\ 
\hline
\textbf{IC}                                                                                                                                          & 0.58                                                                                         & 0.59         & 0.59         & 0.59         & 0.60                      & 0.61           \\ 
\hhline{|=======|}
\begin{tabular}[c]{@{}c@{}}\textbf{Classification}\\\textbf{Accuracy } \end{tabular}                                                                       & 0.45                                                                                         & 0.59         & 0.74         & 0.79         & 0.91                      & 0.98           \\ 
\hline
\begin{tabular}[c]{@{}c@{}}\textbf{Classification}\\\textbf{Accuracy - using}\\\textbf{randomly generated}\\\textbf{initial labelled dataset}\end{tabular} & 0.40                                                                                         & 0.54         & 0.69         & 0.74         & 0.88                      & 0.98           \\
\hhline{|=======|}
\end{tabular}
}
\end{table}

\section{Conclusion}

The manuscript introduces a hybrid query strategy-based framework, which addresses the practical challenges in AL, namely, cold-start, oracle uncertainty and evaluating model performance in the absence of ground truth. The robustness of the proposed framework is evaluated across three datasets from different environments and industrial settings. Firstly, the merits of using a pre-clustering approach to address the cold-start problem is exemplified by experiments performed on the process data collected from a simulated ethanol plant. Secondly, the utility of hybrid query strategies is established with the help of experiments performed on the drug consumption dataset. It is concluded that the use of hybrid query strategies helps in reducing the computational cost, while delivering at par or better performance than the individual query strategies. Finally, the contribution of all the elements together in the proposed framework is demonstrated through experiments carried out on the rolling element bearing test data from Case Western Reserve University. The future extension of this work shall incorporate ways to address other practical challenges in AL, such as variable labelling costs, changing model classes and integrate trust aspects for the model as well as human annotator in the context of explainable artificial intelligence.

\bibliographystyle{ACM-Reference-Format}
\bibliography{sample-manuscript}










\end{document}